\documentclass[pdflatex,referee]{sn-jnl}
\usepackage{graphicx}%
\usepackage{multirow}%
\usepackage{amsmath,amssymb,amsfonts}%
\usepackage{amsthm}%
\usepackage{amsfonts}
\usepackage{lmodern}
\usepackage[title]{appendix}%
\usepackage{xcolor}%
\usepackage{textcomp}%
\usepackage{manyfoot}%
\usepackage{booktabs}%
\usepackage{algorithm}%
\usepackage{algorithmicx}%
\usepackage{algpseudocode}%
\usepackage{listings}%

\theoremstyle{thmstyleone}%
\theoremstyle{thmstyletwo}%

\theoremstyle{thmstylethree}%

\input{preamble}
\usepackage{setspace}
\usepackage{lineno}

\usepackage[authoryear,round]{natbib}
\usepackage{hyperref}       
\hypersetup{
    colorlinks=true,
    citecolor=teal,
    linkcolor=blue,
    urlcolor=cyan,
}

\usepackage{float}

\begin{document}

\title{CamAgent: An LLM-Agent Framework for\vspace{-0.4cm} Multi-Species Camera-Trap Workflows}



\author[1]{\fnm{Yutong} \sur{Deng}}
\equalcont{These authors contributed equally to this work.}

\author[2,3]{\fnm{Qi} \sur{Song}}
\equalcont{These authors contributed equally to this work.}

\author[4]{\fnm{Xi} \sur{Guo}}

\author[2,3]{\fnm{Tianming} \sur{Wang}}

\author*[2,3]{\fnm{Lei} \sur{Bao}}\email{baolei@bnu.edu.cn}

\author*[2,3]{\fnm{Jianping} \sur{Ge}}\email{gejp@bnu.edu.cn}


\affil[1]{\orgdiv{Faculty of Arts and Sciences}, \orgname{Beijing Normal University}, \orgaddress{\city{Zhuhai}, \country{China}}}

\affil[2]{\orgdiv{College of Life Sciences}, \orgname{Beijing Normal University}, \orgaddress{\city{Beijing}, \country{China}}}

\affil[3]{\orgdiv{National Forestry and Grassland Administration Key Laboratory for Conservation Ecology of Northeast Tiger and Leopard}, \orgaddress{\city{Beijing}, \country{China}}}

\affil[4]{\orgdiv{Faculty of Geographical Science}, \orgname{Beijing Normal University}, \orgaddress{\city{Beijing}, \country{China}}}


\abstract{Camera traps accumulated vast, multidimensional data for wildlife monitoring, yet translating raw media archives into meaningful ecological insights remains highly fragmented. Current research workflows require laboriously stitching together disparate analysis tools and scripts, creating steep programming hurdles and complicating end-to-end spatiotemporal analyses. To overcome this fragmentation, we present \textbf{CamAgent}, an autonomous Large Language Model (LLM) agent framework that integrates camera-trap analytical workflows into a unified intelligent ecosystem. CamAgent interprets natural-language ecological intent, schedules computational routing, and executes specialized tools spanning computer-vision perception (e.g., SpeciesNet), CamtrapDP-compatible data management, detection-corrected occupancy modeling, temporal activity analysis, and species co-occurrence networks. The framework automates multi-stage analytical pipelines while maintaining essential data-quality controls and analytical conventions. Consequently, CamAgent significantly reduces manual programming overhead for conservationists, establishing a transparent, scalable, and fully integrated paradigm for camera-trap ecology. Our project is available at \url{https://anonymous.4open.science/r/artifact72c6f4}. }


\maketitle

\section{Introduction}
The widespread deployment of camera traps has fundamentally transformed wildlife monitoring and conservation biology over the past decade~\citep{burton2015wildlife, steenweg2017scaling, mccallum2013changing}. Camera traps extensively deployed across diverse habitats accumulate large-scale, real-world ecological data. Crucially, these monitors provide multidimensional data, combining video sequences with structural metadata like camera identifiers, geographic coordinates, timestamps, species detections, and corresponding model confidence scores~\citep{ahumada2011community, hughey2018challenges,deng2026wild_vlm,song2026boosting}.

Such multidimensional data supports diverse ecological monitoring applications. For instance, integrated timestamps enable researchers to estimate daily activity rhythms and evaluate temporal overlap between interacting species~\citep{ridout2009estimating, rowcliffe2014quantifying}. Geographic coordinates facilitate robust spatial analyses, such as occupancy modeling, habitat use estimation, and biodiversity hotspot identification~\citep{mackenzie2002estimating,tobler2008evaluation}. At the community level, multi-species video sequences within a shared sampling grid enable comprehensive assessment of community composition and complex interspecific relationships, including spatial avoidance and co-occurrence patterns~\citep{ahumada2011community, burton2015wildlife}.

However, the massive accumulation of camera-trap data over the years imposes a severe analytical burden. Transforming these raw data archives into meaningful ecological insights~\citep{tuia2022perspectives, farley2018situating} demands a complex, multi-stage workflow spanning data standardization, deep-learning species recognition, structured database management, and sophisticated statistical modeling~\citep{beery2019efficient, young2018software}. Despite substantial progress in automated perception to enhance wildlife recognition (e.g., deep learning models for species classification and bounding-box detection~\citep{norouzzadeh2018automatically, willi2019identifying,song2024benchmarking}), current methodologies remain \textit{highly fragmented}. The transition from perceptual outputs to structured ecological analysis and visualization still heavily depends on manual processing~\citep{tabak2019machine, niedballa2016camtrapR}. In practice, researchers need to laboriously stitch together disparate recognition pipelines, spreadsheet operations and Python scripts to complete a single ecological workflow.

This fragmentation introduces several critical consequences. Primarily, it imposes steep programming and statistical computing requirements, creating a substantial hurdle for conservationists who seek ecological answers rather than software engineering challenges~\citep{sollmann2018gentle, besson2022towards}. Moreover, separating perception from downstream analysis complicates the creation of coherent end-to-end pipelines, a challenge that is particularly problematic for real-world multi-species datasets requiring joint spatiotemporal interpretation. This disconnect frequently leads to the inconsistent handling of analytical assumptions, such as timestamp reliability, independent-event thresholds, and spatial radii across disconnected scripts. Consequently, verifying the exact relation between final ecological results and the underlying raw field data demands a substantial investment of time and manual labor.

Overcoming this fragmentation requires a system that unifies disparate processing steps and computational tools. The emerging paradigm of Large Language Model (LLM) agents~\citep{yao2022react,wang2024survey} offers a promising solution to this challenge by leveraging advanced capabilities for multi-step reasoning and automated task planning~\citep{schick2023toolformer, wang2024survey, shen2023hugginggpt}. Globally, these agentic frameworks are already driving substantial breakthroughs across diverse empirical sciences, fundamentally shifting traditional research workflows toward autonomous discovery. Recent milestone implementations demonstrate how autonomous LLM agents can independently parse scientific literature, design intricate experimental protocols, and coordinate specialized analytical software and computational pipelines to accelerate complex research workflows~\citep{lu2026towards}. Extending this autonomous scientific assistant paradigm to wildlife monitoring offers a transformative path to handle the expanding computational demands of modern field ecology.

We instantiate this autonomous paradigm through the proposed \textbf{CamAgent}, a framework engineered to systematically process massive multi-species camera-trap archives. Rather than operating as a generative text engine, the central LLM functions strictly as an intelligent workflow coordinator. To deliver specific ecological outcomes, it directly analyzes the user's objective, determines precisely which analytical tools to invoke, and schedules the exact statistical figures or models to be generated to fulfill the request. This query-driven ecosystem consolidates the entire analytical pipeline into a centralized architecture, seamlessly integrating a perception pipeline with detection and classification models~\citep{gadot2024crop,beery2019efficient,redmon2016you}, compatible data management~\citep{bubnicki2024camtrap} and a curated registry of specialized ecological tools. Through a simple natural-language interface, researchers interact directly with the framework, prompting the underlying routing engine to interpret ecological intent, retrieve the necessary spatiotemporal records, parameterize the selected tools, and autonomously synthesize the final results for ecological insights. The main contributions of this paper can be summarized as follows:
\begin{enumerate}
    \item An LLM-agent-based methodological framework is proposed, which organizes camera-trap workflows in a unified, query-driven manner, effectively connecting perception outputs, standardized records, ecological tools, and report-oriented synthesis of results.
    
    \item An analytical pipeline for real-world multi-species data, integrating SpeciesNet-based detection, timestamp normalization, CamtrapDP-compatible exchange, JSON/Chroma-backed storage, conditional filtering, and independent-event construction.
    
    \item A comprehensive suite of ecological tools spanning data quality control, temporal activity analysis, spatial hotspot analysis, detection-corrected occupancy modeling, community composition, species co-occurrence, relationship-network analysis, and visualization.
\end{enumerate}

\noindent Extensive validation demonstrating our LLM-driven framework significantly reduces manual programming effort while explicitly preserving critical computational assumptions, including timestamp reliability, event-independence thresholds, camera effort, and spatial radii.    
\begin{figure}[t]
    \centering
    \includegraphics[width=0.890\linewidth]{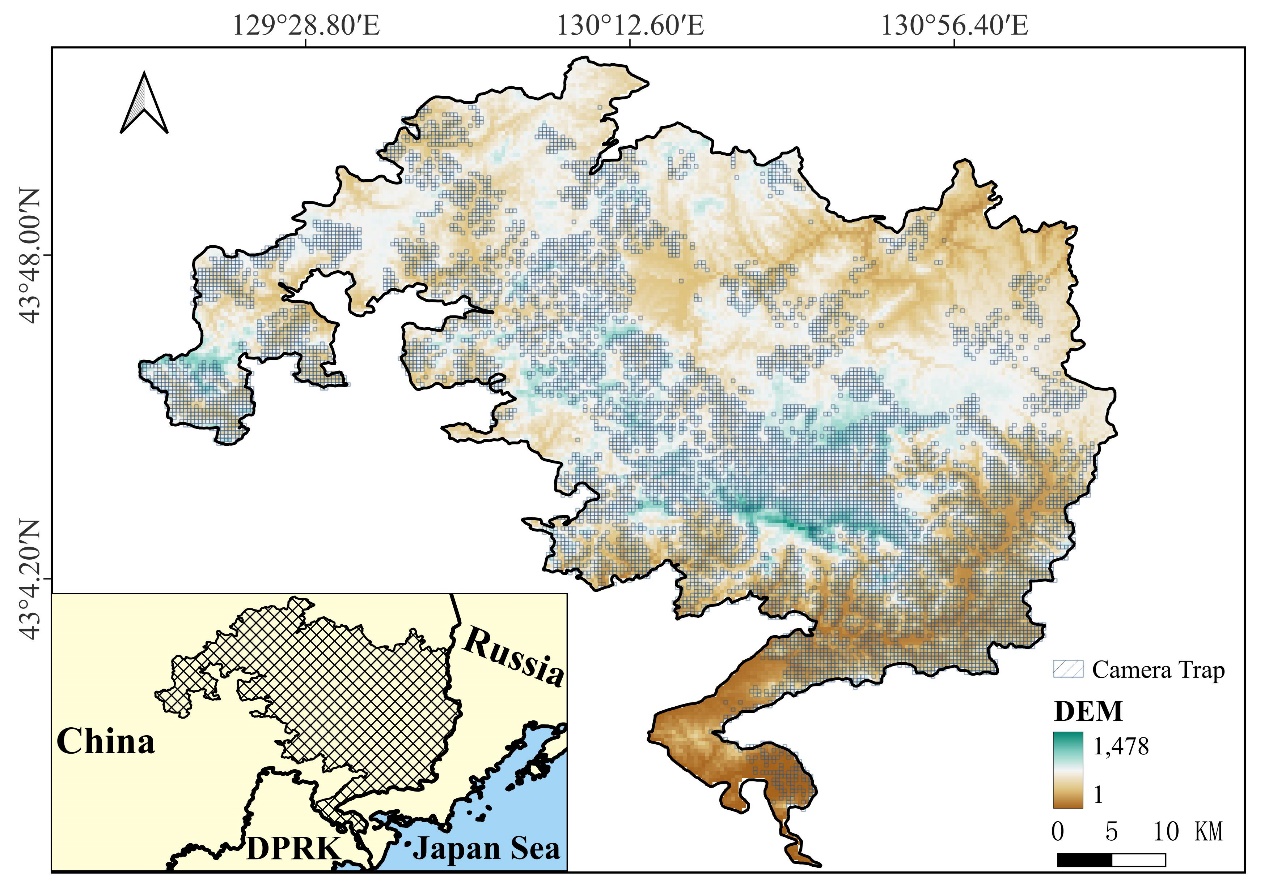}
    \caption{\textbf{Study area and camera-trap sampling footprint in Northeast
China Tiger and Leopard National Park.} The inset locates the study area in northeastern China relative to neighboring countries and the Japan Sea. Hatched grid cells show the camera-trap sampling footprint over a digital elevation model spanning 1--1,478~m. The empirical package used for the six natural-language query experiments
contained 1,376,017 media records from 17,129 camera identifiers between
1 January 2023 and 1 September 2024.}
    \label{fig:camera_traps}
\end{figure}

\section{Materials and Methods}

\subsection{Dataset}
\noindent \textbf{Study Area.}
The primary empirical data are collected from a long-term camera-trap network established across the Northeast China Tiger and Leopard National Park~\citep{wang2016amur}. This park is designed for monitoring endangered Amur tigers and leopards, and the field network provides extensive spatial coverage across the landscape. The camera-trap deployment footprint and elevation gradient are shown in Fig.~\ref{fig:camera_traps}. The bounded analytical package comprised 17,129 camera identifiers and provided repeated non-invasive observations across a large spatial network. 


\noindent \textbf{Data Selection.}
All analyses use records from January 2023 to September 2024, packaged in
CamtrapDP format~\citep{bubnicki2024camtrap}. To check that the pipeline also runs from raw media, we submitted one
camera-trap video with a plain-language request to prepare it for analysis. The planner chose the dataset-preparation
tool, which sampled frames for SpeciesNet~\citep{gadot2024crop} classification,
read the timestamp with PaddleOCR~\citep{du2020pp,cui2025paddleocr},
wrote and validated a CamtrapDP package, and loaded it into the analysis
backend. Five species had at least five independent events (30-min gap) at two or more cameras and were therefore included in the community analysis (Table~\ref{tab:species_dataset}).

\begin{table}[htbp]
    \centering
    \small
    \caption{Media records and independent events for the five species in the
    bounded empirical package. Events were constructed within each
    species--camera combination using a 30-min minimum gap
    (Section~\ref{sec:event_aggregation}). All retained records span January
    2023 to September 2024.}
    \label{tab:species_dataset}
    \begin{tabular}{llrr}
        \toprule
        \textbf{Common name} & \textbf{Scientific name}
            & \textbf{Media records} & \textbf{Events} \\
        \midrule
        Siberian tiger & \textit{Panthera tigris altaica} & 7,745 & 6,570 \\
        Amur leopard & \textit{Panthera pardus orientalis} & 11,704 & 9,882 \\
        Sika deer & \textit{Cervus nippon} & 625,249 & 345,300 \\
        Siberian roe deer & \textit{Capreolus pygargus} & 727,090 & 505,386 \\
        Asiatic black bear & \textit{Ursus thibetanus} & 4,229 & 3,614 \\
        \bottomrule
    \end{tabular}
\end{table}

\subsection{The CamAgent Framework}

CamAgent runs the whole sequence as one workflow. It prepares the raw media, loads the records as CamtrapDP, lets the language model choose which ecological tools to call, runs those tools, draws the figures, and returns an answer alongside the outputs it was built from. 

\begin{figure}[t]
    \centering
    \includegraphics[width=0.99\linewidth]{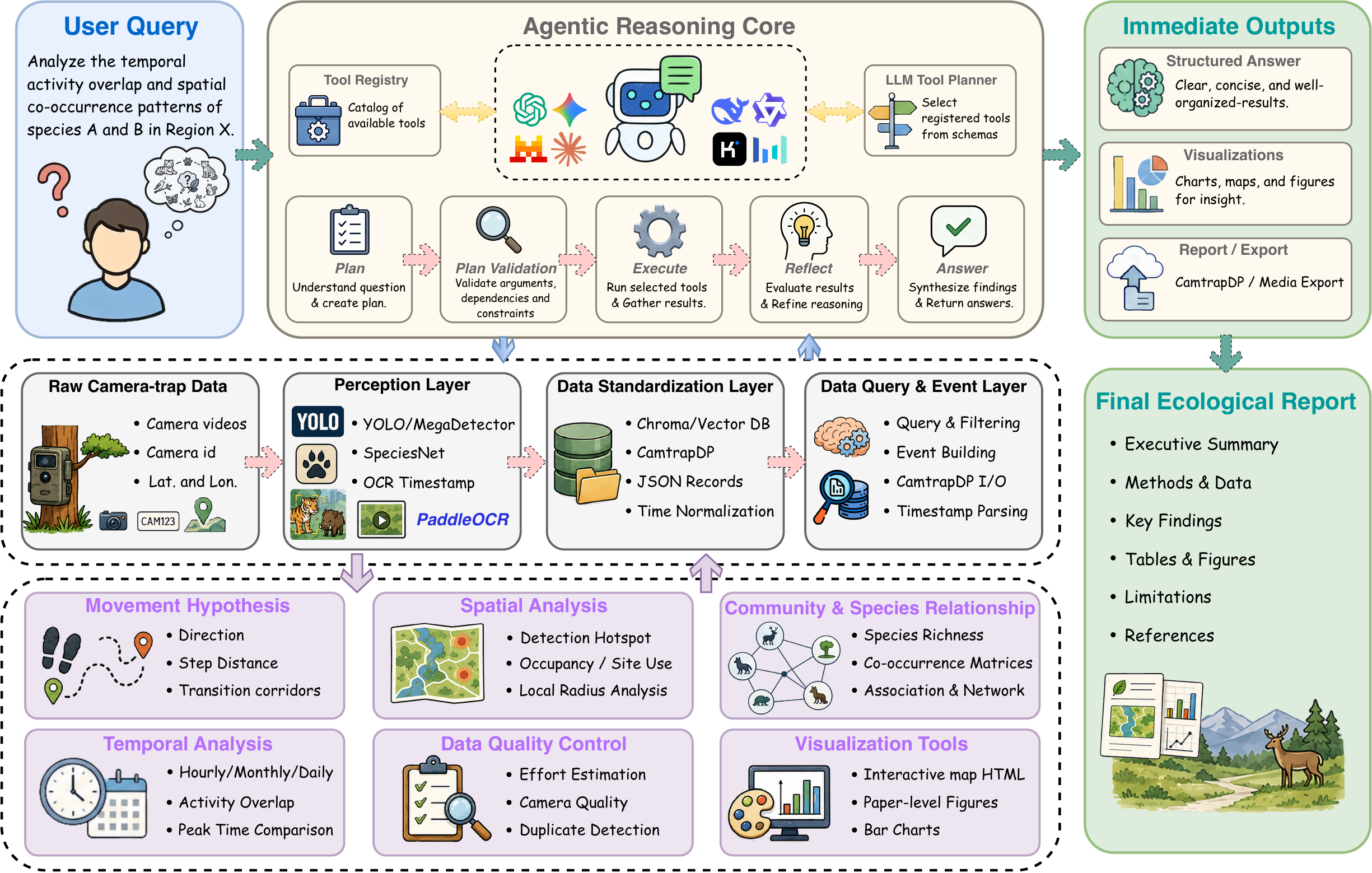}
    \caption{\textbf{The framework of CamAgent.} A user question (left) enters the Agentic Reasoning Core (top), which draws on the Data Layer (middle band) and the Ecological Tool Modules (bottom band) and returns a structured answer, figures, and an exportable report (right). Arrows show the direction of data flow.}
    \label{fig:main}
\end{figure}

CamAgent has three parts (Fig.~\ref{fig:main}). The \textit{Agentic Reasoning
Core} decides what to run for a given question, and it does so in five steps.
It drafts a plan of tool calls from the question and the tool registry,
validates the plan, executes the
calls, reflects on what they returned, and synthesizes an answer. The
\textit{Data Layer} turns raw media into the standardized records those calls
operate on, and the \textit{Ecological Tool Modules} hold the six analysis tools the plan can draw on. The language model performs the planning, reflection and
answer steps, while validation and execution are handled by a local runtime, so
every number and figure in the output comes from a Python function rather than
from the model.

\subsubsection{LLM Planning, Execution, and Reflection}
\begin{figure}[htbp]
    \centering
    \includegraphics[width=1.0\linewidth]{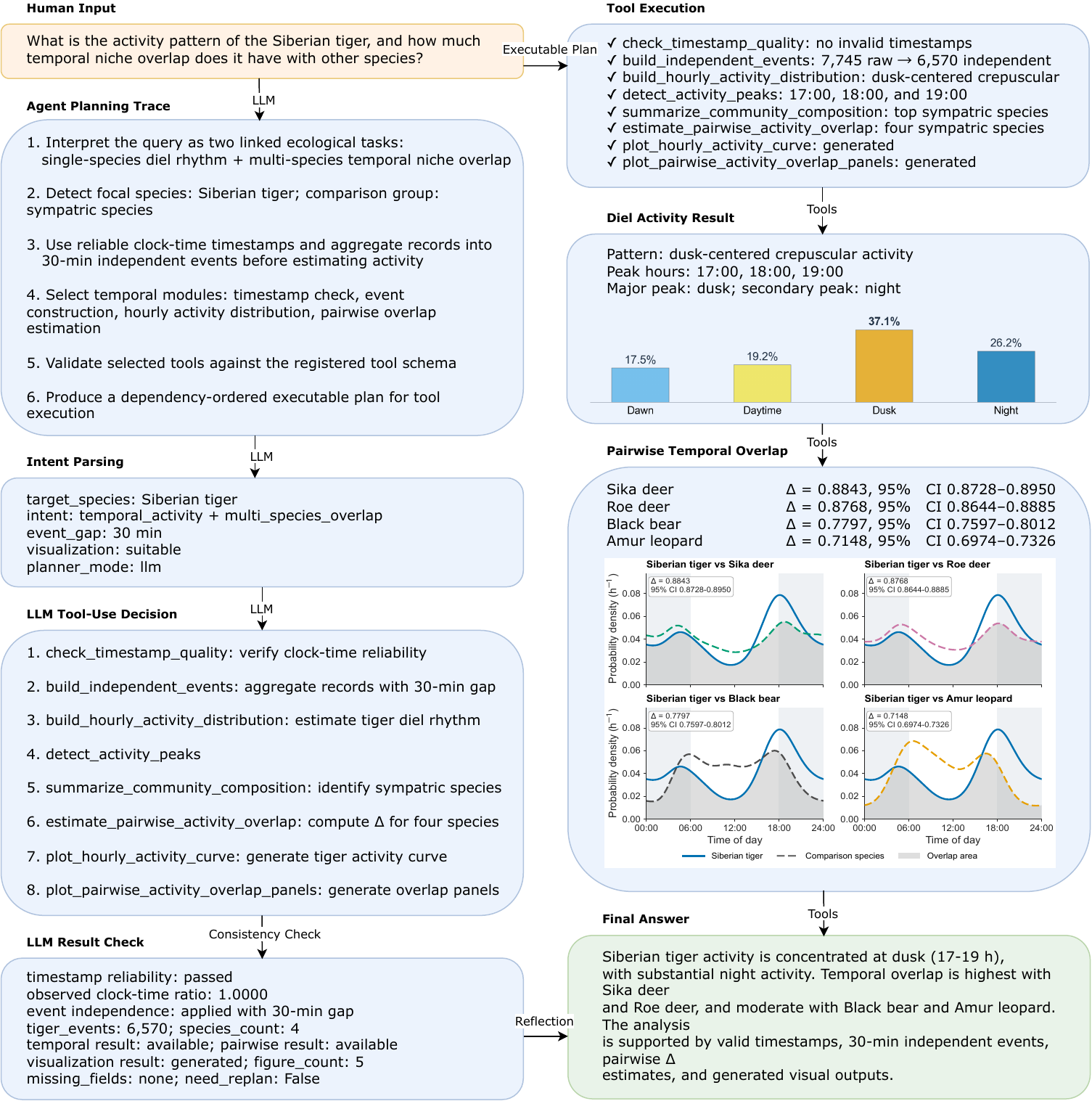}
    \caption{\textbf{The TEMP planning and execution record.} Left: the submitted question, the planning trace, the parsed intent, the eight-call plan, and the reflection check. Right: the execution status of each call, the returned activity and overlap results, and the final answer.}
    \label{fig:agent_trace}
\end{figure}

The agent planner reads the question together with the schemas of the available tools and returns a sequence of tool calls in order. The planning loop follows ReAct~\citep{yao2022react}, with a reflection step added after execution~\citep{shinn2023reflexion}.
Fig.~\ref{fig:agent_trace} shows the trace
for the temporal-activity query (Section~\ref{sec:query_experiments}), in which
the planner returned an eight-call plan, all eight calls succeeded, and five
figures were cited in the answer.

\begin{itemize}
\item \textbf{Tool Registry.} 
Each tool is declared with a name, a JSON
parameter schema, and the preconditions it requires. The planner sees only
these declarations and chooses among them for every question.

\item \textbf{Validation and Execution.} 
Before anything runs, the local runtime
checks tool names, arguments, dependencies, privacy constraints, call budgets,
and each tool's preconditions. It can reject a call, but it will not rewrite the
plan. Calls run in order, and the reflection step inspects what they returned before the answer is drafted. 

\item \textbf{Answer Synthesis.} 
CamAgent assembles the returned tables, maps, and figures into the final answer. The answer reports each result at the level the tool produced it and repeats the constraints of the question set, such as the camera set to be used or the time and distance limits on a movement query.
\end{itemize}

\subsubsection{Data Layer}
The data layer turns raw camera-trap inputs into the standardized records that the analysis tools run on. It accepts images or video together with whatever metadata the field team recorded.

\begin{itemize}
    \item \textbf{Deep Learning Perception.} 
    For images, CamAgent runs SpeciesNet~\citep{gadot2024crop} directly. For video, it samples frames, keeps the per-frame predictions, and aggregates them into one label. PaddleOCR~\citep{du2020pp,cui2025paddleocr} can additionally read a timestamp burned into the frame. The framework also accepts outputs from YOLO~\citep{redmon2016you} or MegaDetector~\citep{beery2019efficient} front-end pipelines when sidecar results are available.

    \item \textbf{Spatiotemporal Standardization.} Camera-trap datasets often arrive with incomplete timestamps, dates written in different formats, and camera coordinates kept in a separate file. CamAgent maps them onto one schema that holds species, timestamp, coordinates, individual count, and a link to the source media. Records that carry only a date are flagged and excluded from the hourly analyses, so only timestamps with a real clock time enter them. We did not convert timestamps to solar time, and therefore all activity results are in local clock time~\citep{frey2017investigating}.

    \item \textbf{Data Management and Interoperability.} Records are stored either as plain JSON files or in a Chroma vector database. CamtrapDP packages can be read into either backend, and processed records exported back out as deployment, media, and observation tables.

    \item \textbf{Querying and Event Aggregation.}
    \label{sec:event_aggregation}
    Records can be filtered by species, date range, confidence threshold, or distance from a point. Consecutive detections of the same species at one camera are collapsed into independent events using a minimum time gap, 30~min by default, to avoid pseudoreplication.

\end{itemize}

\subsubsection{Ecological Tool Modules}
The analysis layer holds six modules. Each reads event records from the data layer and returns tables, figures, and the parameter values it used.

\begin{itemize}
    \item \textbf{Data Quality Analysis.} Checks the records before any analysis runs. It flags missing fields, unparseable timestamps, out-of-range individual counts, and duplicate-like records, and scores each camera on the completeness of its records. 
    
    \item \textbf{Temporal Activity Analysis.} Estimates diel activity from the clock times of independent events by circular von Mises kernel density estimation, and quantifies pairwise overlap with the coefficient of overlapping $\Delta$, the area under the minimum of the two density curves~\citep{ridout2009estimating}. It also reports hourly to seasonal summaries, activity peaks, and the share of events falling in each activity window.
    
    \item \textbf{Spatial Distribution and Guarded Occupancy.} Builds camera-level detection maps, hotspot rankings, and camera-by-occasion detection histories. A single-season occupancy model~\citep{mackenzie2002estimating} is fitted only when documented camera-operational intervals are available, otherwise the tool returns an explanatory error instead of an occupancy estimate.
    
    \item \textbf{Community and Species-Relationship Analysis.} Analyzes every species meeting the minimum-event and minimum-camera thresholds, rather than a fixed focal list. It reports observed and Chao2 richness with species-accumulation curves~\citep{gotelli2001quantifying}, and Hill numbers for alpha and gamma diversity~\citep{chao2014rarefaction}. Beta diversity is summarized with Jaccard and Bray--Curtis dissimilarity, partitioned into S{\o}rensen turnover and nestedness-resultant components~\citep{baselga2010partitioning}. For each species pair, the observed shared-camera count is evaluated under the hypergeometric distribution conditional on the two marginal camera counts~\citep{veech2013probabilistic}, converted to a standardized effect size, and adjusted across pairs with the Benjamini--Hochberg procedure~\citep{benjamini1995controlling}.
    
    \item \textbf{Movement Hypothesis Analysis.} Derives potential transition steps between cameras based on user-defined time-gap and distance constraints. It summarizes distances, directions, and repeated camera pairs. Individual identity is unknown, so a link means two detections were close in time and space, not that one animal moved between the cameras.
    
    \item \textbf{Visualization.} Renders the outputs of the other modules as figures, including temporal activity plots, community composition bars, spatial hotspot maps, detection-history heatmaps, occupancy forest plots, species-relationship panels, and interactive Folium maps.
\end{itemize}

\section{Results}
\subsection{Natural-Language Ecological Question Experiments}
\label{sec:query_experiments}

We ran six natural-language questions through CamAgent, covering data quality (QUAL), temporal ecology (TEMP), spatial ecology (SPAT), community ecology (COMM), movement hypotheses (MOVE), and integrated synthesis (INTEG).
Table~\ref{tab:query_experiments} gives the question and the main output for each experiment. CamAgent completed the six experiments without a failed tool call.

\begin{table*}[htbp]
    \centering
    \small
    \caption{Representative natural-language ecological questions and principal outputs for the six CamAgent experiments.}
    \label{tab:query_experiments}
    \resizebox{\textwidth}{!}{%
    \begin{tabular}{
        p{0.06\textwidth}
        p{0.15\textwidth}
        p{0.48\textwidth}
        p{0.22\textwidth}
    }
        \toprule
        \textbf{Name}
        & \textbf{Question type}
        & \textbf{Query sample}
        & \textbf{Outputs} \\
        \midrule
        QUAL
        & Data-quality assessment
        & Assess the Siberian tiger records for missing fields, invalid timestamps, possible label noise, and duplicate-like records. Explain which downstream ecological analyses remain supported.
        & Required-field completeness, timestamp checks, record-schema flags, duplicate-like groups, analytical restrictions, and Fig.~\ref{fig:qual_summary}. \\
        \midrule
        TEMP
        & Temporal activity and overlap
        & What is the activity pattern of the Siberian tiger, and how much temporal niche overlap does it have with other species?
        & Tiger activity distribution, activity-window shares, pairwise overlap coefficients, uncertainty intervals, and Fig.~\ref{fig:activity_overlap}. \\
        \midrule
        SPAT
        & Spatial distribution and hotspots
        & Where are Siberian tiger detections concentrated across the camera
        network, and which cameras are the main detection hotspots?
        Return a privacy-safe map and do not interpret hotspots as density
        or occupancy.
        & Camera-level hotspot counts, a privacy-masked grid-cell map, detection-only interpretation limits, and Fig.~\ref{fig:spatial_hotspots}. \\
        \midrule
        COMM
        & Dataset-wide community inference
        & Using the current camera-trap data, perform dataset-wide community inference across all eligible species rather than a hard-coded focal list. Estimate observed and Chao2 richness, Hill alpha and gamma diversity, Jaccard and Bray--Curtis beta diversity, partition S{\o}rensen beta diversity into turnover and nestedness components, construct a species-accumulation curve, and test conditional shared-camera associations with false-discovery-rate correction. Return figures.
        & Dataset-wide diversity estimates, a species-accumulation curve, FDR-corrected shared-camera associations, and Fig.~\ref{fig:camera_richness}. \\
        \midrule
        MOVE
        & Camera-detection adjacency hypotheses
        & For Amur leopard detections separated by no more than 72 hours
        and 100 km, summarize candidate inter-camera distances, directions,
        and repeated camera pairs. Treat them as adjacency hypotheses,
        not individual tracks or confirmed corridors.
        & Distance and direction summaries, repeated directed camera-pair counts, interpretation limits, and Fig.~\ref{fig:move_summary}. \\
        \midrule
        INTEG
        & Integrated ecological synthesis
        & Analyze Siberian tiger diel activity and overlap with Sika deer, identify its main detection hotspots, summarize shared-camera
        occurrence between the two species, check the relevant data quality, and return a concise ecological report with figures and limitations.
        & Tiger activity, tiger--sika deer temporal overlap, shared-camera summaries, hotspot counts, and Fig.~\ref{fig:integ_relationship}. \\

        \bottomrule
    \end{tabular}
    }
\end{table*}

\subsection{QUAL: Data-Quality Query Experiment}
The QUAL experiment asked which analyses the tiger records could support.
CamAgent ran a set of checks over the 7,745 records, covering missing fields,
invalid timestamps, suspicious timestamps, schema flags, and duplicate-like groups.

\begin{figure*}[!t]
    \centering
    \includegraphics[width=1.00\linewidth]{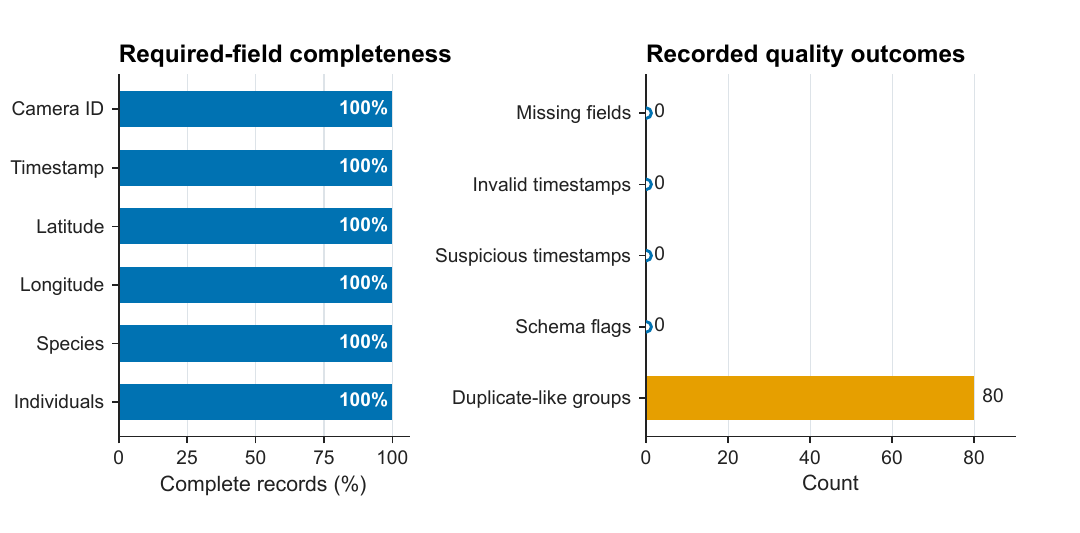}
    \caption{\textbf{Data quality for 7,745 Siberian tiger records.} Left:
completeness of the six required fields. Right: counts of missing fields,
invalid timestamps, suspicious timestamps, schema flags, and duplicate-like
groups.}
    \label{fig:qual_summary}
\end{figure*}

All 7,745 records contained the six required fields and carried a usable clock time, and the schema check flagged nothing. The duplicate check found 80 groups in which several records share a camera, a timestamp and a species. The largest held 202. These are separate media files written by one camera trigger rather than 202 sightings, and the 30-min rule collapses each group into a single event. Temporal and spatial summaries were therefore supported once detections were grouped into events, while occupancy could not
be estimated without survey-effort data (Fig.~\ref{fig:qual_summary}).

\subsection{TEMP: Temporal Activity and Species Overlap}
\begin{figure*}[!t]
    \centering
    \includegraphics[width=1.00\linewidth]{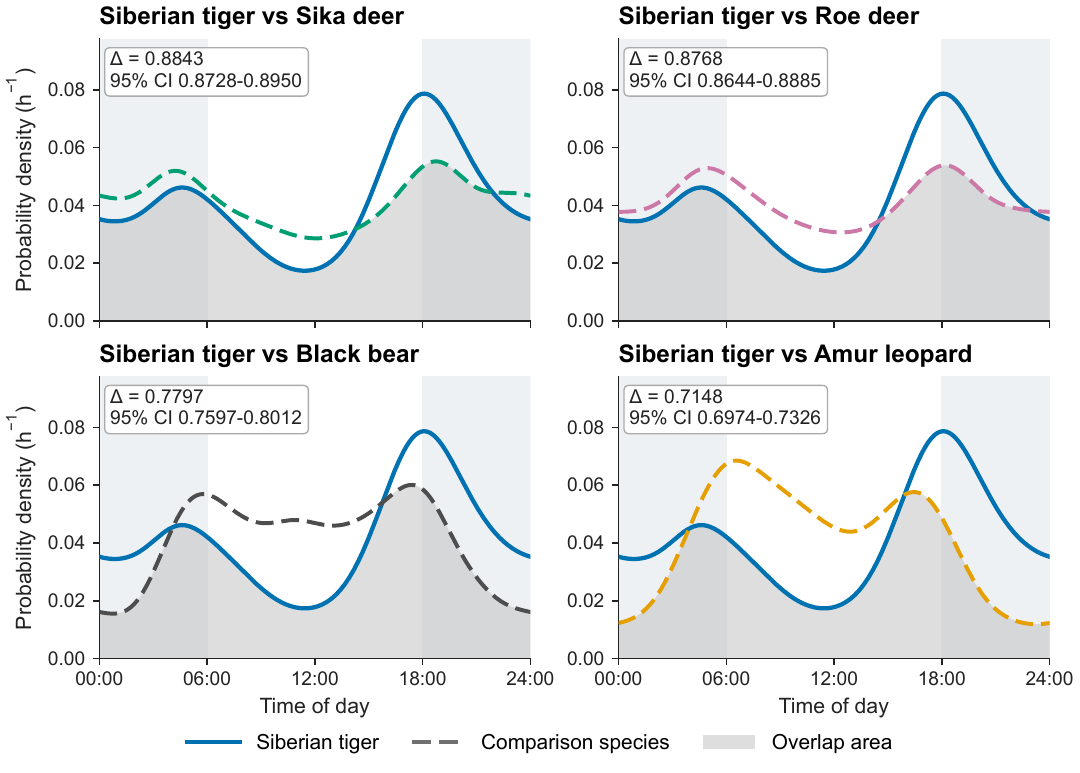}
    \caption{\textbf{Activity overlap between Siberian tiger and four sympatric
species.} Solid blue curves show tiger activity, dashed colored curves the
comparison species, and gray shading the overlap. Sika deer (upper left), roe
deer (upper right), black bear (lower left), Amur leopard (lower right). Each
panel gives $\Delta$ with its 95\% bootstrap interval.}
    \label{fig:activity_overlap}
\end{figure*}

The TEMP question was, \textit{``What is the activity pattern of the Siberian tiger, and how much
temporal niche overlap does it have with other species?''} CamAgent grouped detections into 30-min independent events, estimated hourly activity, and compared tiger activity with sika deer, Siberian roe deer,
Asiatic black bear, and Amur leopard (Fig.~\ref{fig:activity_overlap}).

The 6,570 tiger events peaked at 17:00 to 19:00, with a secondary rise at night (Fig.~\ref{fig:agent_trace}). Most events fell at dusk (37.1\%), then at night (26.2\%), during the day (19.2\%), and at dawn (17.5\%). The four windows are not the same length, so per hour the two peaks are dusk and dawn. Activity overlap was highest with sika deer ($\Delta = 0.8843$, 95\% CI 0.8728--0.8950) and roe deer (0.8768, 0.8644--0.8885), and lower with black bear (0.7797, 0.7597--0.8012) and Amur leopard (0.7148, 0.6974--0.7326)~\citep{ridout2009estimating}.

\subsection{SPAT: Spatial Distribution and Detection Hotspots}
\begin{figure}[!t]
    \centering
    \includegraphics[width=1.00\linewidth]{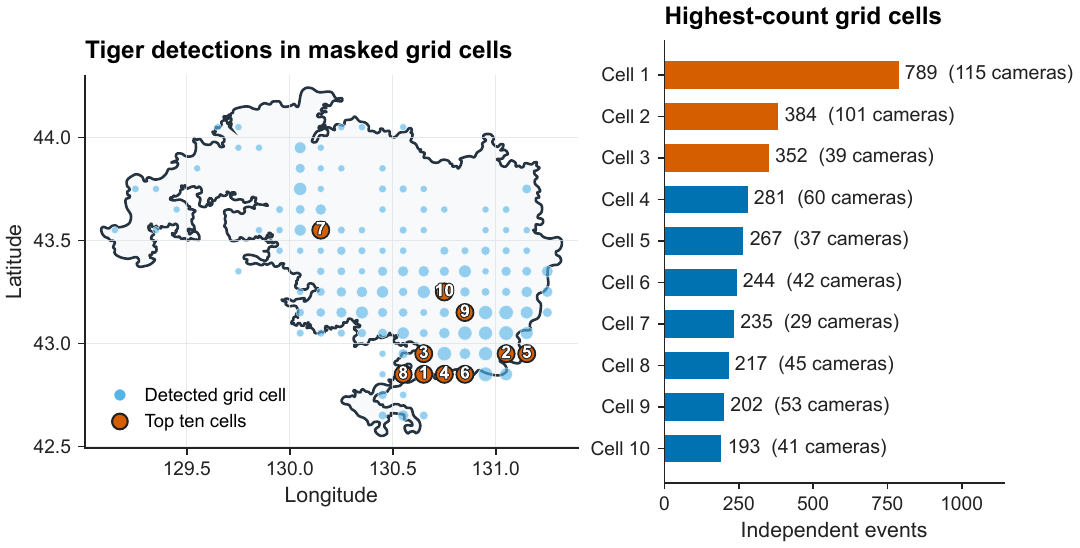}
    \caption{\textbf{Siberian tiger detections in 0.1-degree grid cells.} Numbered
orange symbols mark the ten highest-count cells; the bars give event and camera
counts for each. Camera identifiers are withheld.}
    \label{fig:spatial_hotspots}
\end{figure}

The SPAT experiment asked where tiger detections were concentrated. CamAgent ranked cameras by event count and drew a map with camera locations masked. Tigers were recorded at 1,845 of the 17,129 cameras, and the five highest
camera-level counts were 60, 55, 52, 51, and 50 (Fig.~\ref{fig:spatial_hotspots}).

For public display, events were aggregated to 0.1-degree masked grid cells. Across the ten labeled cells, counts ranged from 193 to 789 events and from 29 to 115 cameras per cell. Camera identifiers were withheld.

\subsection{COMM: Community Composition and Co-occurrence}
For COMM, CamAgent ran diversity estimates, a species-accumulation curve, and shared-camera association tests over all eligible species. All five species passed the thresholds, with 870,752 events across 17,129 cameras. Two ungulates account for 98\% of them (roe deer 505,386; sika deer 345,300). Observed and Chao2 richness were both 5.0~\citep{chao2014rarefaction}. Gamma Hill numbers were 2.210
(95\% CI: 2.198--2.222) for $q=1$ and 2.023
(95\% CI: 2.008--2.037) for $q=2$.

\begin{figure*}[!t]
    \centering
    \includegraphics[width=1.00\linewidth]{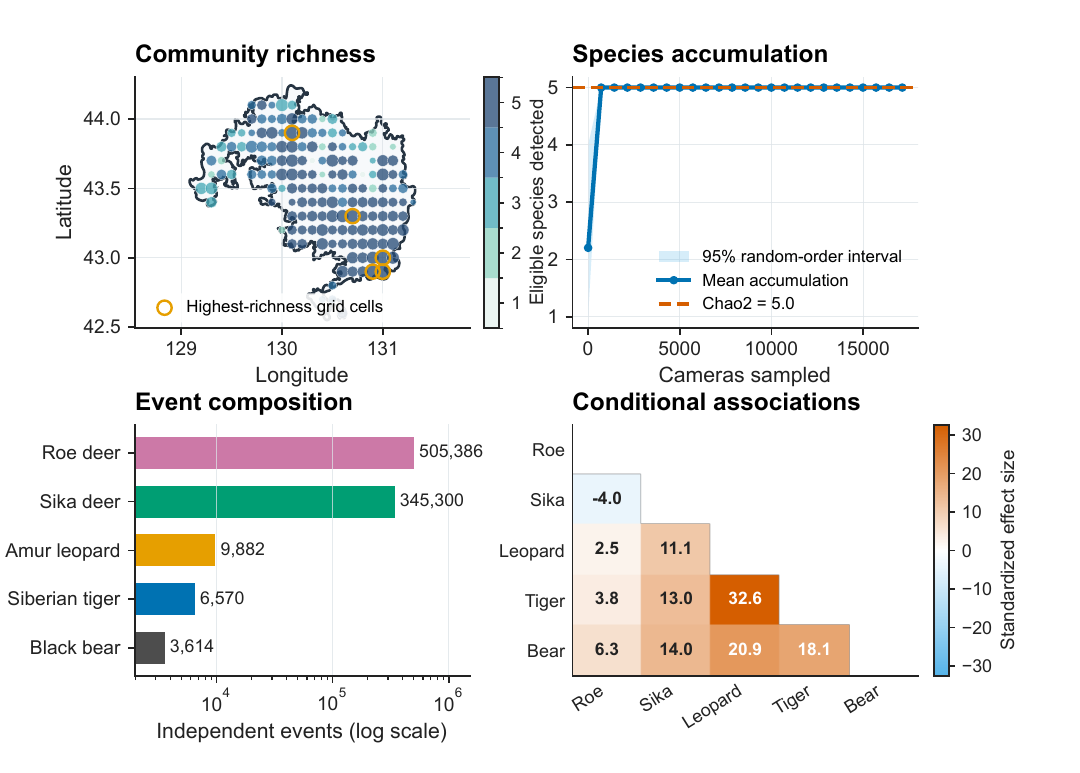}
    \caption{\textbf{Community summary.} Upper left: eligible-species richness by 0.1-degree grid cell. Upper right: species-accumulation curve with observed and Chao2 richness. Lower left: event counts by species. Lower right: standardized effect sizes for shared-camera associations after Benjamini--Hochberg correction.}
    \label{fig:camera_richness}
\end{figure*}

Across 100,000 camera pairs sampled with seed 42, mean Jaccard and Bray--Curtis dissimilarities were 0.320 and 0.634, and mean S{\o}rensen dissimilarity was 0.221, with turnover of 0.040 and a nestedness-resultant component of 0.181~\citep{baselga2010partitioning}. With 17,129 cameras all ten tests were significant after Benjamini--Hochberg correction~\citep{benjamini1995controlling}, so we read the effect sizes rather than the p-values. Nine effects were positive, one
(roe deer--sika deer) was negative, and the largest standardized effect was
observed for Amur leopard--Siberian tiger (32.63) (Fig.~\ref{fig:camera_richness}).

\subsection{MOVE: Camera-Detection Adjacency Hypotheses}
\label{sec:movement}

\begin{figure*}[!t]
    \centering
    \includegraphics[width=1.00\linewidth]{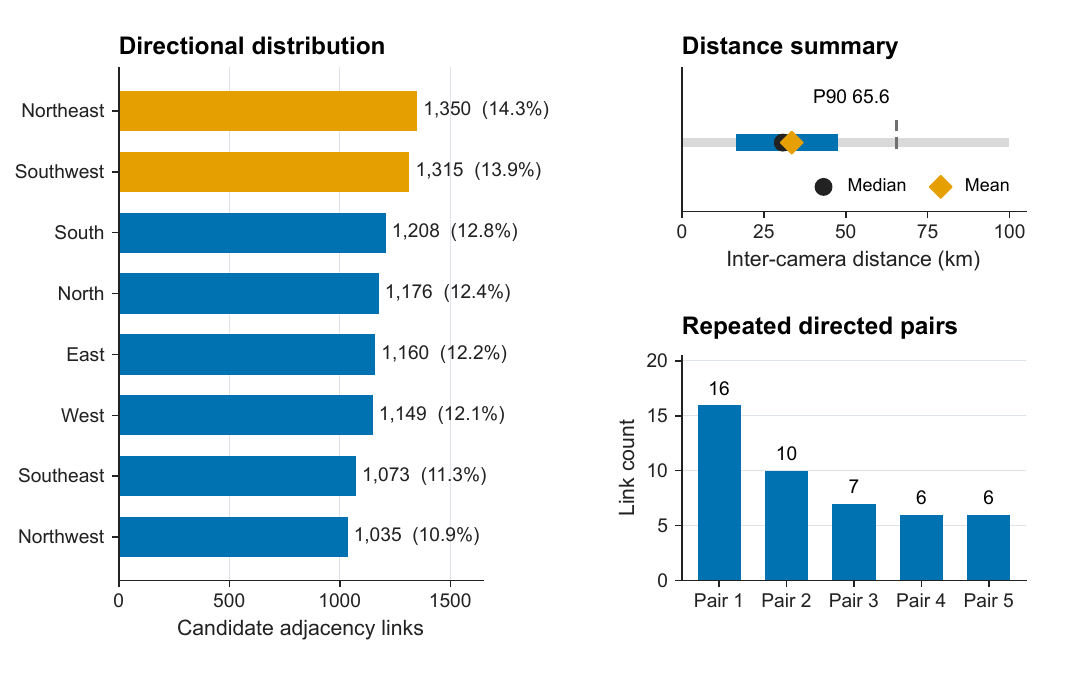}
    \caption{\textbf{Amur leopard links within 72~h and 100~km ($n=9{,}466$).} Left: link counts across eight directional bins. Upper right: distance distribution with median, mean, and 90th percentile. Lower right: the five most frequent directed camera pairs.}
    \label{fig:move_summary}
\end{figure*}

The MOVE experiment asked CamAgent to link Amur leopard detections falling within 72 hours and 100 km of each other. It found 9,466 such links and summarized their distances, directions, and repeated camera pairs (Fig.~\ref{fig:move_summary}). Links spanned a median of 30.8~km (IQR 16.5--47.7~km).

Directions were close to uniform across the eight bins (12--14\%), with no dominant axis. No camera pair recurred often enough to suggest a route, the most frequent appearing 16 times among 9,466 links. Individual identity is unknown, so a link means two detections were close in time and space, not that one animal moved between the two cameras~\citep{kays2015terrestrial}.

\subsection{INTEG: Combining Analyses in One Query}

One request chained data-quality checking, tiger activity, tiger--sika deer overlap, detection hotspots, shared-camera occurrence, visualization, and report writing. It used 6,570 tiger and 345,300 sika deer events.

\begin{figure*}[!t]
    \centering
    \includegraphics[width=1.00\linewidth]{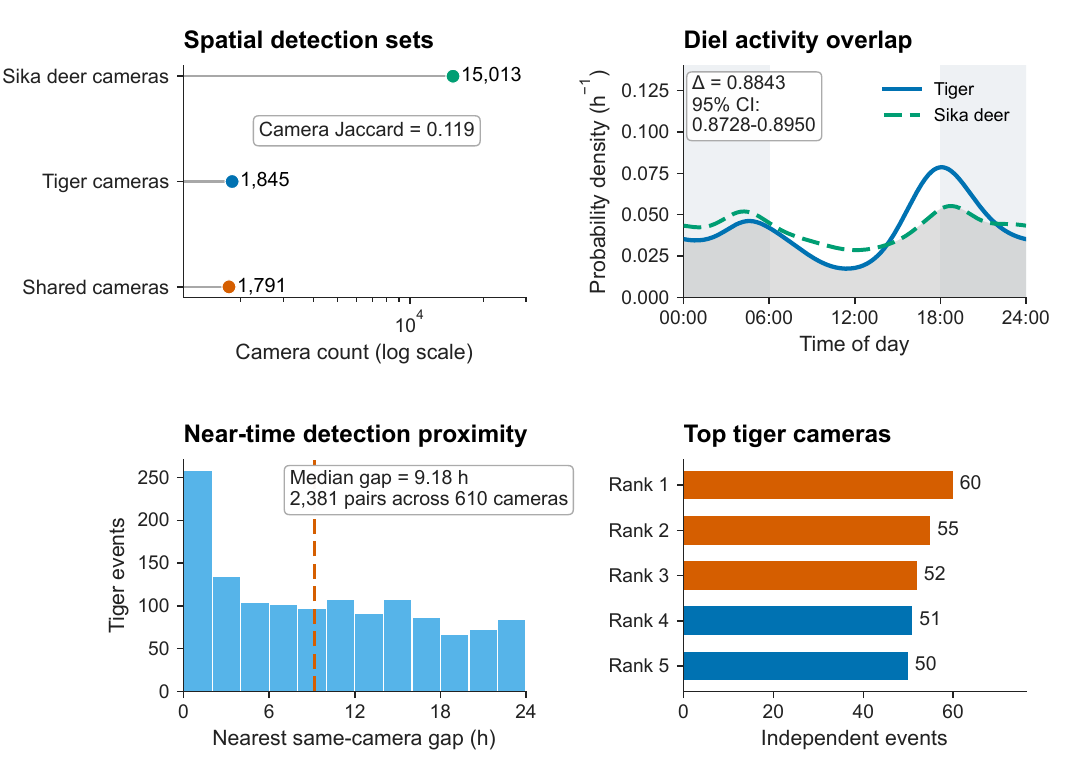}
    \caption{\textbf{Siberian tiger and sika deer.} Upper left: camera sets for each species and their overlap. Upper right: activity overlap (same estimate as Fig.~\ref{fig:activity_overlap}). Lower left: nearest same-camera cross-species gaps within 24~h. Lower right: the five highest tiger event counts by camera, identifiers withheld.}
    \label{fig:integ_relationship}
\end{figure*}

Overlap with sika deer was $\Delta = 0.8843$ (95\% CI 0.8728--0.8950), the same
estimate as in TEMP. Of the 1,845 cameras that recorded tigers, 1,791 also
recorded sika deer. At 610 of those, 2,381 cross-species event pairs fell
within 24~h, with a median gap of 9.2~h (Fig.~\ref{fig:integ_relationship}).

\section{Discussion}
The primary contribution of \textbf{CamAgent} lies in resolving the acute workflow fragmentation that has long bottlenecked camera-trap data analysis. Traditional research workflows rely heavily on disjointed software stacks—forcing ecologists to manually transfer outputs between computer-vision models, spreadsheet operations, and disconnected scripts or packages such as \texttt{camtrapR}~\citep{sollmann2018gentle, young2018software, tabak2019machine, niedballa2016camtrapR, besson2022towards}. Transforming these vast, raw media archives into meaningful ecological insights creates steep programming barriers and undermines end-to-end analytical reproducibility~\citep{tuia2022perspectives, farley2018situating}. In contrast, CamAgent integrates multi-stage analytical processes—from perception and standardized exchange formats like CamtrapDP~\citep{bubnicki2024camtrap} to downstream spatiotemporal statistical modeling—into a unified, query-driven ecosystem. By organizing multi-dimensional observational streams (comprising spatial coordinates, temporal timestamps, video sequences, and species detections) into a cohesive analytical continuum, CamAgent operationalizes the concept of \textit{geo-ecological data cubes}, enabling direct natural-language querying for scalable biodiversity monitoring~\citep{burton2015wildlife, steenweg2017scaling}.

Unlike recent fully autonomous ``AI Scientists'' implementations that independently generate hypotheses, execute experiments, and write manuscripts end-to-end~\citep{lu2026towards}, CamAgent adopts a strategically controlled, hybrid design philosophy. Generative language models are inherently susceptible to hallucinations and numerical inaccuracies when performing statistical calculations directly. To overcome this limitation, CamAgent strictly separates high-level task planning from deterministic scientific computation~\citep{schick2023toolformer, yao2022react}. The central LLM operates strictly as an intelligent workflow coordinator, retrieving validated tools from a schema-defined registry and passing parameters to execute deterministic code execution engines~\citep{wang2024survey, shen2023hugginggpt}. Upstream perception outputs from deep learning detectors~\citep{beery2019efficient, norouzzadeh2018automatically, willi2019identifying} are seamlessly fed into downstream statistical modules. This separation guarantees complete traceability: critical analytical rules and assumptions—such as timestamp verification, independent-event interval thresholds, and spatial buffer radii~\citep{ridout2009estimating, rowcliffe2014quantifying, mackenzie2002estimating}—remain fully transparent, customizable, and reproducible.

This query-driven architecture demonstrates exceptional scalability when transitioning from single-focal-species assessments to complex community-level evaluations. When prompted with multi-species community queries, CamAgent dynamically orchestrates multi-tier analytical pipelines without requiring hard-coded target species lists. It seamlessly coordinates Hill diversity partitioning (alpha, beta, and gamma components), rarefaction-extrapolation curves, and false discovery rate (FDR) corrected co-occurrence matrix evaluations~\citep{ahumada2011community, burton2015wildlife, gotelli2001quantifying, chao2014rarefaction, baselga2010partitioning, benjamini1995controlling}. Crucially, while CamAgent automates these complex multi-species computations, it serves as a computational accelerator rather than an automated ecological interpreter. For instance, shared camera encounters or overlapping activity rhythms identified by the framework provide empirical spatiotemporal patterns, but do not inherently constitute proof of direct interspecific behavioral avoidance or interaction~\citep{dou2019prey, xiao2018relationships}. The agentic system delivers rigorous, standardized statistical foundations, leaving final ecological inferences to domain specialists.

Our work provides an extensible foundational framework for integrating camera-trap archives into broader multidimensional geo-ecological data cubes. While the current framework seamlessly processes standardized metadata formats and raw media files, future extensions can incorporate multi-modal vision-language models capable of extracting fine-grained behavioral sequence labels directly from video streams~\citep{beery2018recognition, alencar2026advancing, deng2026wild_vlm, song2026boosting}. Furthermore, coupling this intelligent workflow agent with continuous spatial environmental rasters—including high-resolution digital elevation models, land cover, microclimate, and remote sensing time series~\citep{kays2015terrestrial}—will enable autonomous, end-to-end habitat suitability and animal movement dynamics modeling. By unifying perception, data standardization, and complex spatiotemporal modeling under a natural-language interface, CamAgent overcomes key computational bottlenecks~\citep{hughey2018challenges}, establishing a scalable, transparent paradigm for automated ecosystem status assessment and long-term biodiversity monitoring.
\section{Conclusion}
Camera-trap archives have outgrown the workflows used to analyze them. We introduced \textbf{CamAgent}, an LLM agent that converts natural-language questions into explicit, re-runnable tool plans for registered camera-trap tools. In six natural-language queries the model chose its own tools for data quality, activity, space, community, and movement questions. Every call it proposed passed validation and ran, and the six runs produced 11 figures. Numbers come from Python functions, not from the model. CamAgent takes camera-trap media and a plain-language question and returns the analyses, the figures, and a report, with a record of how each is produced.
\section*{Author Contribution}
Yutong Deng: Conceptualization, Methodology,  Writing - original draft. Qi Song: Conceptualization, Methodology, Writing - original draft. Xi Guo: Visualization, Writing - review \& editing. Tianming Wang: Supervision, Funding acquisition, Writing - review \& editing. Lei Bao: Supervision, Funding acquisition,  Writing - review \& editing. Jianping Ge: Supervision, Funding acquisition, Resources, Project administration, Writing - review \& editing.

\section*{Data Availability Statement}
The source code and data sample are available at \url{https://anonymous.4open.science/r/artifact72c6f4}. The full camera coordinates are not allowed to be publicly distributed, as they contain sensitive border information.

\section*{Funding}
This research is funded by the National Key Research and Development Program of China (grant number 2024YFF1307301).

\section*{Declaration of Generative AI and AI-assisted Technologies}
During the preparation of this work, the author(s) used Google Gemini in order to refine language, improve readability, and perform language editing. After using this tool, the author(s) thoroughly reviewed and edited the content as needed and take responsibility for the content of the publication.

\bibliographystyle{elsarticle-harv}
\bibliography{reference}

\end{document}